\documentclass{article}

\usepackage[preprint]{neurips_2025}
\usepackage{amsmath}
\usepackage{graphicx}
\usepackage{multirow}
\usepackage{array}

\usepackage[utf8]{inputenc} 
\usepackage[T1]{fontenc}    
\usepackage{hyperref}       
\usepackage{url}            
\usepackage{booktabs}       
\usepackage{amsfonts}       
\usepackage{nicefrac}       
\usepackage{microtype}      
\usepackage{xcolor}         

\title{Deep Learning Surrogates for Real-Time Gas Emission Inversion}

\author{
Thomas Newman \\
School of Mathematical Sciences\\
Lancaster University\\
Lancaster, LA1 4YF \\
\texttt{t.newman1@lancaster.ac.uk} \\
\And
Christopher Nemeth \\
School of Mathematical Sciences\\
Lancaster University\\
Lancaster, LA1 4YF \\
\texttt{c.nemeth@lancaster.ac.uk} \\      
\AND
Matthew Jones \\
Shell Global Solutions International BV\\
Amsterdam, 1031 HW \\
\texttt{matthew.m.jones2@shell.com} \\
\And
Philip Jonathan\\
School of Mathematical Sciences\\
Lancaster University\\
Lancaster, LA1 4YF \\
\texttt{p.jonathan@lancaster.ac.uk} \\
}

\begin{document}

\maketitle

\begin{abstract}
Real-time identification and quantification of greenhouse-gas emissions under transient atmospheric conditions is a critical challenge in environmental monitoring. We introduce a spatio-temporal inversion framework that embeds a deep‐learning surrogate of computational fluid dynamics (CFD) within a sequential Monte Carlo algorithm to perform Bayesian inference of both emission rate and source location in dynamic flow fields. By substituting costly numerical solvers with a multilayer perceptron trained on high-fidelity CFD outputs, our surrogate captures spatial heterogeneity and temporal evolution of gas dispersion, while delivering near-real-time predictions. Validation on the Chilbolton methane release dataset demonstrates comparable accuracy to full CFD solvers and Gaussian plume models, yet achieves orders-of-magnitude faster runtimes. Further experiments under simulated obstructed-flow scenarios confirm robustness in complex environments. This work reconciles physical fidelity with computational feasibility, offering a scalable solution for industrial emissions monitoring and other time-sensitive spatio-temporal inversion tasks in environmental and scientific modeling.
\end{abstract}


\section{Introduction}\label{intro}

Methane, the principal constituent of natural gas, is a potent greenhouse gas with a global warming potential over 80 times greater than that of carbon dioxide over a 20-year period \citep{stocker2014climate}. Despite its relatively short atmospheric lifetime compared to carbon dioxide, methane is highly effective at absorbing infrared radiation, making it a critical driver of near-term climate warming. Consequently, reducing emissions of methane is considered one of the
most effective near-term actions to limit climate change.

Methane emissions occur in various natural environments such as wetlands, but the main contribution to emissions are anthropogenic sources in sectors like agriculture waste, biomass burning and fossil fuel production and use \citep{saunois2025global}. In oil and gas production, methane emissions typically come from flaring, venting, leaks or incomplete combustion. Accurate identification and quantification of these sources can help the development of effective policies, regulations, and target remediation. The Oil and Gas Methane Partnership 2.0 (OGMP2.0) initiative exemplifies industry efforts to standardize monitoring and reporting of these emissions \citep{OGMP2024}. Because multiple emission sources can be present at a facility, some of them varying in time, sensing technologies are needed that deliver full spatio-temporal resolution.

Recovering unknown source parameters from sensor data requires inversion techniques. In this work, we focus on ground-level sensor networks delivering high-frequency, spatially sparse measurements. A Bayesian inversion framework – traditionally relying on Markov chain Monte Carlo (MCMC) \citep{brooks2011handbook,robert1999monte,fearnhead2024scalable} – quantifies uncertainty robustly but suffers from poor scalability and slow convergence in multimodal, time-varying settings. Real-time inference necessitates an alternative that balances accuracy and efficiency.

\paragraph{Our Contribution:} We embed a computational fluid dynamics (CFD) surrogate within a Bayesian state-space model \citep{kitagawa1996monte,kalman1960new} and perform sequential inference using a Sequential Importance Resampling (SIR) particle filter \citep{gordon1993novel}. Our surrogate employs a multilayer perceptron (MLP) trained to emulate high-fidelity CFD solvers, providing near-instantaneous predictions of sensor concentrations for any candidate source configuration. This approach retains the physical realism of numerical solvers while reducing per-evaluation cost to milliseconds.

In this paper, we apply our inversion methodology to the Chilbolton controlled-release dataset \citep{hirst2018mapping, hirst2020methane, voss2024multi, newman2024probabilistic}, demonstrating that our surrogate-based SIR filter achieves comparable accuracy to full CFD and Gaussian plume models \citep{stockie2011mathematics} at a fraction of the computational cost. We further validate robustness under simulated obstructed flow fields and temporally varying emission rates, highlighting the scalability of our framework. Section~\ref{section2} presents the SIR-based inversion algorithm. Section~\ref{section3} details the MLP surrogate construction. Section~\ref{section4} evaluates performance on Chilbolton data, and Section~\ref{section5} extends the approach to complex, obstructed scenarios.

\section{Spatio-temporal gas source inversion using particle filters}
\label{section2}

We cast the gas source inversion problem in a Bayesian state-space framework \citep{jazwinski1970stochastic, kitagawa1996monte, douc2005comparison,west2006bayesian}, which is commonly used in a range of areas, including econometrics \citep{aoki1991state,hamilton1994state}, target tracking \citep{arulampalam2002tutorial,nemeth2012particle,stone2013bayesian} and epidemiology \citep{PRASHAD2025591,hooker2011parameterizing}. Let the latent state at time $t$ be $\boldsymbol{\theta}_t$, which encodes the unknown source parameters – notably the source coordinates $(\tilde{x}, \tilde{y}, \tilde{z})$ and possibly a time-varying emission rate $s_t$. We specify a prior distribution $p(\boldsymbol{\theta}_0)$ over the initial state (for example, a uniform prior over the site for the source location and a broad prior for the emission rate) to capture our initial uncertainty. The state-space model is then defined by two components: a \textit{measurement model}, which relates the state to the observed gas concentrations, and a \textit{state evolution model} describing how the latent state changes over time.

\paragraph{Measurement model (Observation Equation).} At any time $t$, we receive sensor measurements $d_t$ (e.g. gas concentration readings at fixed sensor locations). We model these observations as noisy functions of the current source parameters. In particular, we assume the observation equation:

\begin{equation}\label{observation prediction equation}
    \hat{\boldsymbol{d}}_t = C(\dot{\boldsymbol{x}}, \dot{\boldsymbol{y}}, \dot{\boldsymbol{z}} \mid \Tilde{x}, \Tilde{y}, \Tilde{z}) \times \boldsymbol{s}_{\kappa:t} + \boldsymbol{\beta}_{\kappa:t} + \boldsymbol{\epsilon}_t,
\end{equation}

where $(\dot{x},\dot{y},\dot{z})$ are the known coordinates of the sensor location(s), and $C(\dot{x}, \dot{y}, \dot{z}\mid\tilde{x},\tilde{y},\tilde{z})$ is the gas concentration function (derived from a CFD model or its surrogate) that predicts the concentration at the sensors given a source at $(\tilde{x}, \tilde{y}, \tilde{z})$. The term $\boldsymbol{s}_{\kappa:t}$ denotes the history of the source’s emission rate from time $t-\kappa$ up to $t$, representing the fact that gas concentration at time $t$ can depend on emissions in the recent past (within a window of length $\kappa$). Likewise, $\boldsymbol{\beta}_{\kappa:t}$ is the history of the ambient background gas concentration at the sensor (e.g. baseline methane levels) over that period. The final term $\boldsymbol{\epsilon}_t$ represents measurement noise (sensor error), which we typically model as independent zero-mean Gaussian noise. If multiple sensors are deployed, $\boldsymbol{d}_t$ is a vector of all sensor readings at time $t$, and we assume the components of $\boldsymbol{\epsilon}_t$ are independent (i.e., each sensor has independent noise). For simplicity, we also assume the source’s influence on the flow field is negligible (i.e. the wind field is not altered by the emission), so that the dispersion of gas is linearly related to the emission rate. This means $C(\cdot)$ can be computed for a unit emission and then scaled by $\boldsymbol{s}_{\kappa:t}$, as reflected in \eqref{observation prediction equation}.

\paragraph{State evolution model (Dynamic Equation).} We allow the source parameters to evolve in time according to a latent dynamics model. In general, the source location might be static (or moving slowly), and the emission rate $s_t$ could potentially vary over time. We capture any uncertainty or evolution in these parameters with a stochastic dynamic equation. A simple choice (which we adopt here) is a random-walk model:

\begin{equation}\label{state equation}
\boldsymbol{\theta}_{t} = \boldsymbol{\theta}_{t-1} + \boldsymbol{\zeta}_{t},
\end{equation}

where $\boldsymbol{\zeta}_t \sim \mathcal{N}(0,\mathbf{W})$ is a multivariate Gaussian process noise term with covariance $\mathbf{W}$. This model assumes that from one time step to the next, the source location and emission rate do not change dramatically, only undergoing small random perturbations. In practice, if the true source is stationary (constant location and emission), the random walk (with a small covariance $\mathbf{W}$) serves to maintain diversity in our particle filter (preventing all particles from collapsing to a single point). If the source emission rate genuinely varies over time, $\mathbf{W}$ can be tuned to account for those changes. We assume the state process ${\boldsymbol{\theta}}_t$ is \textit{Markovian} (i.e. given $\boldsymbol{\theta}_{t-1}$, the next state $\boldsymbol{\theta}_t$ is independent of earlier times) and that process noise $\boldsymbol{\zeta}_t$ is independent across time steps. We also assume observations are conditionally independent given the corresponding state (with the caveat that $\boldsymbol{d}_t$ may depend on a short history $\boldsymbol{s}_{\kappa:t}$ of emissions, which can be incorporated by extending the state to include recent emissions up to $\kappa$). Under these assumptions, the model is a state-space model amenable to Bayesian filtering techniques to recover the latent process $\boldsymbol{\theta}_{1:t}$.

\paragraph{Bayesian filtering.} Our goal is to infer the posterior distribution of the source parameters given the sequence of measurements up to time $t$, denoted $p(\boldsymbol{\theta}_t \mid \boldsymbol{d}_{1:t})$. Using Bayes’ rule, and the state-space model assumptions, the posterior can be updated sequentially: starting from a prior $p(\boldsymbol{\theta}_0)$, we incorporate new observations as they arrive. In principle, the update from time $t-1$ to $t$ is given by:

\begin{itemize}
    \item \textbf{Prediction:} $p(\boldsymbol{\theta}_t \mid \boldsymbol{d}_{1:t-1}) = \int p(\boldsymbol{\theta}_t \mid \boldsymbol{\theta}_{t-1}) p(\boldsymbol{\theta}_{t-1} \mid \boldsymbol{d}_{1:t-1}) \text{d}\boldsymbol{\theta}_{t-1}$,
    \item \textbf{Bayes Update:} $p(\boldsymbol{\theta}_t \mid \boldsymbol{d}_{1:t}) \propto p(\boldsymbol{d}_t \mid \boldsymbol{\theta}_t) p(\boldsymbol{\theta}_t \mid \boldsymbol{d}_{1:t-1}).$
\end{itemize}

Here, $p(\boldsymbol{d}_t \mid \boldsymbol{\theta}_t)$ is the likelihood of the new observation given state $\boldsymbol{\theta}_t$, which from \eqref{observation prediction equation} (assuming Gaussian sensor noise) can be written as, for example, $p(\boldsymbol{d}_t \mid \boldsymbol{\theta}_t) = \mathcal{N}\big(\boldsymbol{d}_t \big| C(\dot{x},\dot{y},\dot{z}\mid\tilde{x},\tilde{y},\tilde{z}) \times \boldsymbol{s}_{\kappa:t} + \boldsymbol{\beta}_{\kappa:t}, \sigma^2 \big)$ for some noise variance $\sigma^2$. In general, these integrals and proportionality are intractable to solve in closed form due to the nonlinearity of $C$ and the high dimensionality of the state. We therefore resort to a Monte Carlo approximation -- specifically, the Sequential Importance Resampling (SIR) particle filter \citep{gordon1993novel} -- to perform the Bayesian update numerically.

In a SIR particle filter, we maintain a set of $N$ random samples (particles) $\{\boldsymbol{\theta}_t^{(i)}\}_{i=1}^N$ that provides a discrete approximation of the posterior $p(\boldsymbol{\theta}_t \mid \boldsymbol{d}_{1:t})$. Each particle $\boldsymbol{\theta}_t^{(i)} = \{\tilde{x}^{(i)}, \tilde{y}^{(i)}, \tilde{z}^{(i)}, \boldsymbol{s}^{(i)}\}$ is a possible source location and emission rate. We also associate a weight $w_t^{(i)}$ with each particle, indicating its relative plausibility given the data. The particle filter sequentially propagates and updates these weighted samples as new data arrive:

\begin{enumerate}
    \item \textbf{Initialization:} At $t=0$, draw $N$ particles $\{\boldsymbol{\theta}_0^{(i)}\}$ from the prior $p(\boldsymbol{\theta}_0)$, and set all weights $w_0^{(i)} = 1/N$.
    \item \textbf{Prediction (Propagate):} For each particle at time $t-1$, sample a new particle according to the state equation \eqref{state equation}. In practice, we add independent Gaussian noise to each particle -- $\boldsymbol{\theta}_t^{(i)} \sim p(\boldsymbol{\theta}_t \mid \boldsymbol{\theta}_{t-1}^{(i)})$. This yields a predicted particle set $\{\boldsymbol{\theta}_t^{(i)}\}_{i=1}^N$ representing an approximate prior for time $t$.
    \item \textbf{Update (Weight):} Upon receiving the observation $d_t$, compute a likelihood weight for each particle based on how well that particle’s state explains the measurement. For particle $i$: 
    $$w_t^{(i)} \propto w_{t-1}^{(i)} p\big(\boldsymbol{d}_t \mid \boldsymbol{\theta}_t^{(i)}\big),$$
using the observation model \eqref{observation prediction equation}.
\end{enumerate}

In practice, the particles are often resampled, with probability $w_t^{(i)}$, to avoid weight degeneracy -- where one particle carries most of the weight. After resampling, all weights are reset to $w_t^{(i)}=1/N$, for all $i$. 

We repeat the Prediction–Update–Resampling cycle for each time step as new sensor data become available. Over time, the particle ensemble $\{\boldsymbol{\theta}_t^{(i)}, w_t^{(i)}\}_{i=1}^N$ evolves to track the posterior distribution $p(\boldsymbol{\theta}_t \mid \boldsymbol{d}_{1:t})$. In effect, the particle filter provides a numerical approximation of the Bayesian solution for the gas source inversion problem. This approach is well-suited to our setting as it can handle nonlinear and non-Gaussian relationships (unlike, e.g., a Kalman filter \citep{kalman1960new}) and it naturally accommodates the sequential arrival of data in a dynamic environment. By using a sufficiently large number of particles, the SIR filter can approximate the true posterior to any desired accuracy, enabling robust spatio-temporal estimation of the source location and emission rate even under complex, unsteady flow conditions.

\section{Multilayer perceptron surrogate modeling of
atmospheric gas measurements in unsteady-state flow fields}\label{section3}

The particle filter outlined in Section \ref{section2} requires repeated evaluations of the likelihood $p(\boldsymbol{d}_{1:t} \mid \boldsymbol{\theta})$ for each particle $\{\boldsymbol{\theta}_t^{(i)}\}_{i=1}^N$, which in the case of the gas concentration model, requires computing the concentration function $C(\dot{x},\dot{y},\dot{z}\mid\tilde{x},\tilde{y},\tilde{z})$ at each potential source location. High-fidelity CFD models exist to compute $C$, but they are often too slow to run for each particle $i$ and time step $t$ in real-time. We can overcome this bottleneck by utilizing a deep learning surrogate model for the CFD simulation. Specifically, we train a multilayer perceptron (MLP) \citep{rosenblatt1958perceptron} to approximate the mapping from the source parameters to the sensor measurements, effectively serving as a fast emulator of the physical gas dispersion model. 

\subsection{High-fidelity gas dispersion simulation training data}

To generate training data for the surrogate, we first require a \textit{ground-truth} model of how gas disperses in an unsteady flow field. We use a computational fluid dynamics (CFD) solver \citep{holl2024phiflow} that captures the physics of air flow and gas transport in our monitored site $\Omega$. In particular, we solve the time-dependent Navier-Stokes equations \citep{navier1821lois, naiver1827lois} (which governs fluid flow) to obtain the wind velocity field, and then solve the advection-diffusion equation \citep{smoluchowski1916brownsche,chandrasekhar1943stochastic} (which governs the transport and diffusion of the gas) to obtain gas concentrations. These equations are discretized and integrated over time to simulate the evolution of wind and gas in the domain. Denoting by $f_v$ the Navier-Stokes solver and $f_c$ the advection-diffusion solver, we can formalize the process as follows: given a history of wind boundary conditions (e.g. measured wind speed and direction over time) $u_{\kappa:t}$ and corresponding pressure field $p_{\kappa:t}$, and given any fixed obstacles or terrain features $\omega$ over the spatial domain $\Omega$, the CFD model produces a \textit{flow field} $f_v(u_{\kappa:t},p_{\kappa:t},\omega,\Omega)$ describing the wind velocities in $\Omega$ over time. Using the flow field, the gas transport solver $f_c$ computes the resulting gas concentration field for a source at a specific location $(\tilde{x},\tilde{y},\tilde{z}).$ We then evaluate this concentration field at the sensor coordinates $(\dot{x},\dot{y},\dot{z})$. Let $C_\text{ns}$ denote the concentration output of the full Navier-Stokes-based numerical solver. We can express the solver's prediction as:

\begin{equation}\label{numericalsolvereq} C_{\text{ns}}(\dot{x}, \dot{y}, \dot{z} \mid \Tilde{x}, \Tilde{y}, \Tilde{z}) = 
f_c((\Tilde{x}, \Tilde{y}, \Tilde{z}), f_v(\boldsymbol{u}_{\kappa:t}, \boldsymbol{p}_{\kappa:t}, \omega, \Omega))\Big\rvert_{\dot{x}, \dot{y}, \dot{z}},
\end{equation}
which represents the gas concentration at the sensor location due to a source at $(\tilde{x},\tilde{y},\tilde{z})$ under the given unsteady wind conditions. In other words, we use the CFD solver to simulate the propagation of gas from a candidate source through the evolving wind field, and we record what concentration would be measured at the sensor. Equation \ref{numericalsolvereq} is essentially the model behind the concentration function $C$ in \eqref{observation prediction equation}. This high-fidelity simulation accounts for complex effects, such as turbulent eddies and time varying wind direction, providing accurate \textit{ground truth} concentrations for given source parameters. 

However, running such a CFD simulation for every candidate source is computationally expensive. For example, solving the Navier-Stokes and advection-diffusion equations even once (for a given source configuration) might take seconds to hours, which is prohibitively expensive when deployed within a particle filter that could require thousands of evaluations. Therefore, we will use \eqref{numericalsolvereq} offline to generate training datasets, and then train a fast MLP surrogate to mimic its output. 

To construct the training data, we sample a large number of hypothetical source scenarios and simulate each with the CFD model. In our study, we drew source locations uniformly from the area of interest $\Omega$ (each location $(\tilde{x},\tilde{y}, \tilde{z})$ corresponds to a different training sample). For each source location, we assume a fixed emission rate (e.g. a unit emission for simplicity) and run the CFD solver to obtain $C_{\text{ns}}(\dot{x},\dot{y},\dot{z}\mid\tilde{x},\tilde{y},\tilde{z})$ via \eqref{numericalsolvereq}. All simulations use the same physical environment and flow conditions representative of the scenario we care about. In particular, we leverage data from the Chilbolton experiment (see Section \ref{section4}): a time series of wind measurements (from an anemometer) provides the unsteady wind boundary conditions $u_{\kappa:t}$ for the solver, and the site is relatively flat and unobstructed (no large $\omega$ features), which allows us to simplify the simulation. Because the vertical variation in this experiment was minimal, we perform the CFD simulations in two dimensions (assuming all sources and sensors lie in the same horizontal plane). This two-dimensional approximation greatly reduces computational cost while introducing only small errors for a flat site. We use the recorded time-varying wind profile uniformly across the domain (spatially uniform but temporally varying wind) when solving the Navier–Stokes equations, given the small size of the site. In summary, our dataset consists of many pairs $\{(\tilde{x}^{(i)}, \tilde{y}^{(i)}), \boldsymbol{d}^{(i)}\}$, where $(\tilde{x}^{(i)},\tilde{y}^{(i)})$ is a sampled source location and $\boldsymbol{d}^{(i)} = C_{\text{ns}} (\dot{x},\dot{y}\mid\tilde{x}^{(i)},\tilde{y}^{(i)})$ is the corresponding sensor reading produced by the CFD simulation (for a given wind sequence and unit emission). These synthetic data samples form the ground truth that our MLP will learn to emulate.
 
\subsection{MLP surrogate model: architecture and training}

We design a multilayer perceptron (MLP) to serve as a surrogate for the CFD-based concentration function. The MLP is a fully-connected feed-forward neural network that takes the source location as input and outputs the predicted gas concentration at the sensor. In our case, the input vector to the MLP is $(\tilde{x},\tilde{y})$ (the two-dimensional coordinates of a potential source). The output is the predicted sensor measurement $\boldsymbol{d}$ (or a vector of concentrations if multiple sensors are present – one output per sensor). Because the relationship from source location to sensor concentration can be quite complex (highly nonlinear due to the physics of dispersion), we choose a sufficiently expressive network architecture. In our implementation, for example, we use an MLP with several hidden layers (e.g. 4 - 8 layers) and each hidden layer has on the order of hundreds of neurons. We employ SeLU activations \citep{klambauer2017self} at the hidden layers (SeLU: Scaled Exponential Linear Unit, designed to self-normalize the neural network while avoiding exploding/vanishing gradients and dying neurons), and a linear activation at the output layer (since we are performing a regression to predict a continuous concentration value). We initialize the network weights using standard Xavier initialization \citep{glorot2010understanding} and train them to minimize the error between the MLP’s predictions and the true concentrations from the CFD simulations.

\paragraph{Training.} We use a supervised learning approach to train the MLP on the dataset of simulated source scenarios. We define a loss function $\mathcal{L}$ as the Mean Squared Error (MSE) between the MLP’s prediction $C_{\text{MLP}}$ and the ground-truth solver output $C_{\text{ns}}$ over all training samples. Formally, if the training set is $\{(\tilde{x}^{(i)}, \tilde{y}^{(i)}, \boldsymbol{d}^{(i)})\}_{i=1}^N$ (with $\boldsymbol{d}^{(i)} = C_{\text{ns}}(\dot{x},\dot{y}\mid\tilde{x}^{(i)},\tilde{y}^{(i)})$, as above), the loss is:

\[
\mathcal{L}(\boldsymbol{\Theta}) = \frac{1}{N}\sum_{i=1}^N \Big\|C_{\text{MLP}}\big(\tilde{x}^{(i)}, \tilde{y}^{(i)}; \boldsymbol{\Theta}\big)-\boldsymbol{d}^{(i)}\Big\|^2,
\]
where $\boldsymbol{\Theta}$ denotes the learnable weights of the network. We minimize this loss using stochastic gradient descent. The training is run for enough epochs until the error plateaus without overfitting, which in our experiments was on the order of a few tens of thousands of epochs. After training, we obtain an approximate functional mapping $C_{\text{MLP}}$ that is a fast proxy for the CFD solver’s output $C_{\text{ns}}$, meaning the MLP’s prediction of sensor measurements for a given source is nearly the same as the high-fidelity CFD prediction \eqref{numericalsolvereq}, but can be computed instantaneously. Once this surrogate is trained, it can be plugged into the particle filter: whenever we need to evaluate the likelihood of a particle (i.e. compute the expected sensor reading for a potential source), we use the MLP instead of running a CFD simulation.

\paragraph{Practical Considerations.} The flow conditions are time-varying, so the mapping from source location to sensor reading could drift over long periods as winds change. To handle the temporal non-stationarity, we adopt a sliding time-window approach in training the surrogate. Instead of training a single MLP on the entire duration of data (which might force it to average over different wind regimes), we train separate MLP models for consecutive time-windows of the data. For example, we can segment the simulation (and real data) into shorter intervals (each spanning a few minutes), and train one MLP on data from each interval. By keeping the time-window short – on the order of the gas transport time across the site – we ensure each MLP sees a relatively homogeneous wind condition, allowing it to more accurately learn the input-output mapping for that period. In effect, the surrogate model is updated periodically to account for changes in the flow. In our case, the window length can be chosen based on the maximum travel time for gas to reach the farthest sensor (which depends on wind speed and domain size). When deploying the inversion in real-time, the particle filter can then switch to the appropriate MLP corresponding to the current time-window of data. This sequential MLP training strategy enables the surrogate to capture transient behaviors and temporal evolution of the gas plume that a single global model might miss.

\paragraph{Post-Training Assessment.} We evaluate the MLP surrogate on held-out test cases (source locations not seen during training) to ensure it generalizes well. We use metrics such as the mean absolute percentage error (MAPE) between $C_{\text{MLP}}$ and $C_{\text{ns}}$ on these test simulations, and we also compare the surrogate’s outputs against real sensor measurements when available (e.g. from the Chilbolton release trial). The MLP consistently achieves low prediction error, indicating that it captures the physical relationship between source and sensor effectively. Furthermore, the surrogate is extremely fast: evaluating the MLP for a given input takes on the order of milliseconds or less, which is orders of magnitude faster than running a full CFD simulation for the same scenario. In fact, our learned model is even faster than the simplified Gaussian plume equations \citep{stockie2011mathematics} (which are themselves a closed-form approximation) while retaining the accuracy of the CFD approach. This balance of physical fidelity and computational efficiency is what enables our overall inversion framework to operate in near-real-time. In the next section, we demonstrate that using the MLP surrogate within the particle filter yields accurate and rapid source inversion results, effectively combining rigorous Bayesian estimation with a fast learned physics model.

\section{Real-data: Chilbolton gas emission inversion}
\label{section4}

We now assess the performance of our inversion framework on real data from the Chilbolton Observatory, where controlled methane releases were conducted under varying atmospheric conditions. The known ground truth for source locations and emission rates provides a basis for quantitative validation.

The dataset includes path-averaged methane concentration measurements collected using a laser dispersion spectrometer, which scanned seven retroreflectors every 3 seconds. Wind measurements were obtained via a three-dimensional ultrasonic anemometer and the sources consisted of perforated $2\text{m} \times 2\text{m}$ ground frames; see Supplementary Materials for site layout. The flat topography and high-frequency measurements make this dataset well suited for evaluating spatio-temporal inversion methods.

\subsection{Sensor measurements prediction}

We first evaluate the predictive performance of the MLP surrogate model against two baselines: a traditional Gaussian plume model and the high-fidelity numerical solver used for training data generation. The Gaussian plume model is a widely adopted, closed-form analytical solution to the advection-diffusion equation; often used for its computational efficiency. Its solution describes a steady-state atmospheric gas transport corresponding to a long-term averaged transport in unobstructed spatially uniform wind fields. Predictions from our three models were made minute-by-minute and evaluated against the last 20-second averaged sensor measurements. Predicting the last 20 seconds is consistent with the time-averaging Gaussian plume model assumption and ensures the numerical solver's simulated gas reaches the sensors from anywhere on the Chilbolton site. For each minute interval, the Gaussian plume model used the averaged wind inputs, while the numerical solver simulated transport using the full minute of wind data and averaged the last 20 seconds.

The MLP was trained to predict the numerical solver's 20-second averaged sensor measurements using 484 simulations with distinct source locations, all under the same wind boundary conditions obtained by our anemometer. The true source location was held out to assess interpolation performance.  All models were then given true source locations and evaluated on two test cases: 10 minutes of data from Source 1 (release 2) and 15 minutes from Source 2 (release 5) -- these reflect periods of ideal wind conditions, where sensors were exposed to the gas plumes.

Table \ref{MAPE} reports the MAPE for each model. The numerical solver achieved the lowest MAPE but required approximately 21 minutes of computational time for all predictions. The Gaussian plume model was much faster (1.2 seconds) but substantially less accurate. The MLP surrogate achieved accuracy close to the numerical solver, outperforming the plume model, while requiring only milliseconds per prediction -- faster than even the plume model. It took 6 minutes to generate the data used to train each MLP and 2 minutes for model training using 90 CPUs and 250GB memory, though this can be reduced with GPU acceleration. Each MLP comprises 4 hidden layers with 100 neurons per layer. These results confirm that the MLP surrogate delivers both high accuracy and real-time prediction capability in unsteady flow conditions.

\begin{table}
\caption{Gaussian plume model, numerical solver and surrogate model predictions' MAPE for 10 minutes of Source 1 release 2 and 15 minutes of Source 2 release 5 given true source locations. Computational times highlight the extreme efficiency of the surrogate.}
\vspace{2mm}
\label{MAPE}
\centering
\begin{tabular}{llll}
    \hline
        & Gaussian Plume  & Numerical Solver & MLP \\
        \hline
        Source 1 & 17.25\% & 13.28\% & 13.38\%  \\
        \hline
        Source 2 & 10.63\% & 9.19\% & 9.57\% \\
        \hline
        Time (s) & 1.20 & 1220.53 & 0.25  \\
        \hline
\end{tabular}
\end{table}

\subsection{Gas source inversion}

We next evaluate the inversion framework by estimating source location using the MLP surrogate within the SIR particle filter. As a baseline, we compare to an inversion using an atmospheric stability class-free Gaussian plume model following \cite{newman2024probabilistic}, reducing model misspecification introduced by traditional Gaussian plume models. To ensure informative updates, we used 4-minute sliding time- windows of data over the 10 minutes of Source 1 measurements -- therefore using four MLPs and one Gaussian plume model per window.

\begin{figure}
\centering
  \includegraphics[width=.5\linewidth]{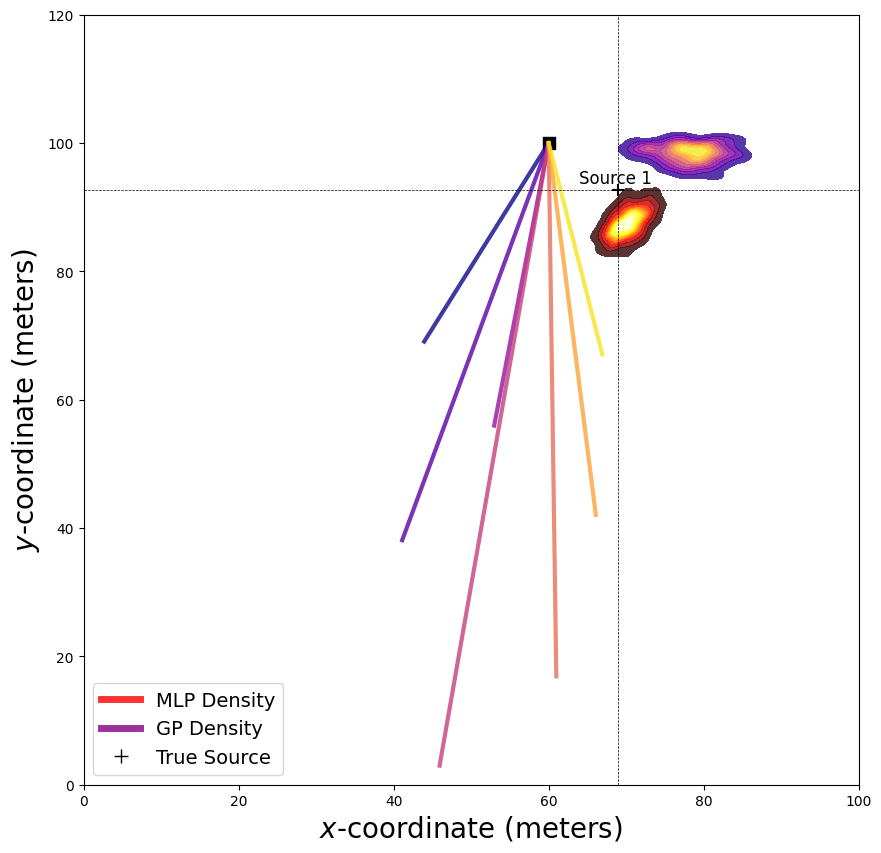}
  \caption{Gaussian plume model and surrogate model-based SIR particles' posterior density of Source 1's location using 700 iterations and 1,000 particles. The posterior density closer to the true source location represents the surrogate-based inversion providing more accurate parameter estimation. }
\label{source1inversionplot}
\end{figure}

\begin{table}
\caption{Gaussian plume model and surrogate model-based SIR particles' mean estimation of Source 1's location. The mean is computed by averaging the distance from all particles at the SIR last iteration. Computational times highlight the efficiency of the surrogate -- the MLP computational time includes training data generation, training, and SIR inversion.}
\label{source1inversiontable}
\centering
\begin{tabular}{llll}
\hline
\multicolumn{2}{l}{} & \multicolumn{1}{l}{MLP} & \multicolumn{1}{l}{Gaussian Plume} \\
\hline
\multirow{2}{*}{Source 1} & Distance (m) & 5.82 & 11.09 \\
                           & Time (min) & 83.3 & 173.6 \\
\hline
\end{tabular}
\end{table}

Figure \ref{source1inversionplot} shows the posterior distribution of estimated source locations for Source 1 using both the MLP and plume-based filters -- the MLP-based inversion yields tighter and more accurate posterior. Table \ref{source1inversiontable} quantifies this, reporting the mean distance across all particles for the last SIR iteration. The MLP-based inversion reduced mean localization error by nearly half compared to the plume model (5.82m vs. 11.09m) and required only half of the computation time (83.3 min vs. 173.6 min), including surrogate training and particle filtering.
Together, these results demonstrate that our surrogate-based framework achieves high inversion accuracy with substantially reduced computational cost, enabling real-time spatio-temporal inference with quantified uncertainty in real-world scenarios.


\section{Case-study: source inversion in obstructed unsteady-state flow fields}\label{section5}

We now demonstrate the scalability and robustness of our proposed inversion methodology in synthetically generated, more complex monitoring environments -- specifically, those featuring obstacles and time-varying emissions, for which no real datasets are currently available. We simulate three distinct 10-minute methane emission events with temporally fluctuating emission rates, each within a spatial domain populated by obstructions. Detailed simulation parameters and setup are provided in the Supplementary Materials.

To emulate real-world operational constraints, we implement a sequential inversion protocol in which sensor data are processed minute-by-minute and a 3-minute sliding time-window of the data is used for the likelihood evaluation. A new MLP surrogate is trained each minute on data from the most recent flow conditions, and the SIR particle filter is subsequently updated by sliding the data window, refining the posterior over source parameters. Specifically, 499 CFD-based training simulations are used to train each MLP, with training performed in parallel across 90 CPU cores, consuming approximately 250 GB of memory. The inversion itself -- including 100 iterations of particle filtering between each minute with 1,000 particles -- is executed on a modest workstation with only 4 CPU cores and 15 GB of memory. Each MLP comprises four hidden layers with 500 neurons per layer; architectural and training details are further elaborated in the Supplementary Materials.

\begin{figure}
\centering
  \includegraphics[width=.8\linewidth]{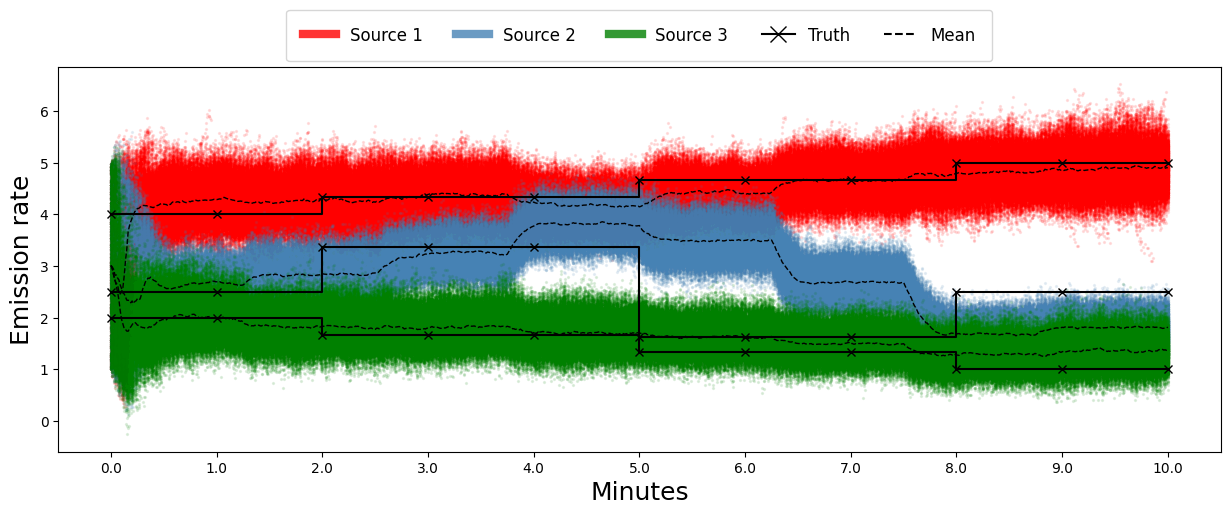}
  \caption{Case study results for temporally varying emission rate estimation. The surrogate-based SIR successfully tracks increasing, decreasing and fluctuating emission rates. The delayed adjustment after Source 2's emission rate shape drop at minute 5 highlights a limitation -- such scenarios may require a more responsive two-layered SIR filter.}
\label{casestudyrates}
\end{figure}

Figure \ref{casestudylocation} visualizes the posterior over source locations after the full 10-minute observation window, clearly demonstrating the model’s ability to accurately infer fixed source positions -- even when occluded by structural obstacles. Figure \ref{casestudyrates} further highlights the framework’s capability to track dynamically varying emission rates: all three sources exhibit time-varying emission profiles, and our particle-based posterior adapts accordingly, accurately reconstructing the temporal evolution of each source’s emission intensity.  However, the delayed adjustment in estimating Source 2’s emission rate following a sharp drop at minute 5 reveals a limitation of the SIR filter -- abrupt changes in emission behavior may require more responsive approaches, such as an interacting multiple model filter \citep{blom2002interacting}.

\begin{figure}
\centering
  \includegraphics[width=.5\linewidth]{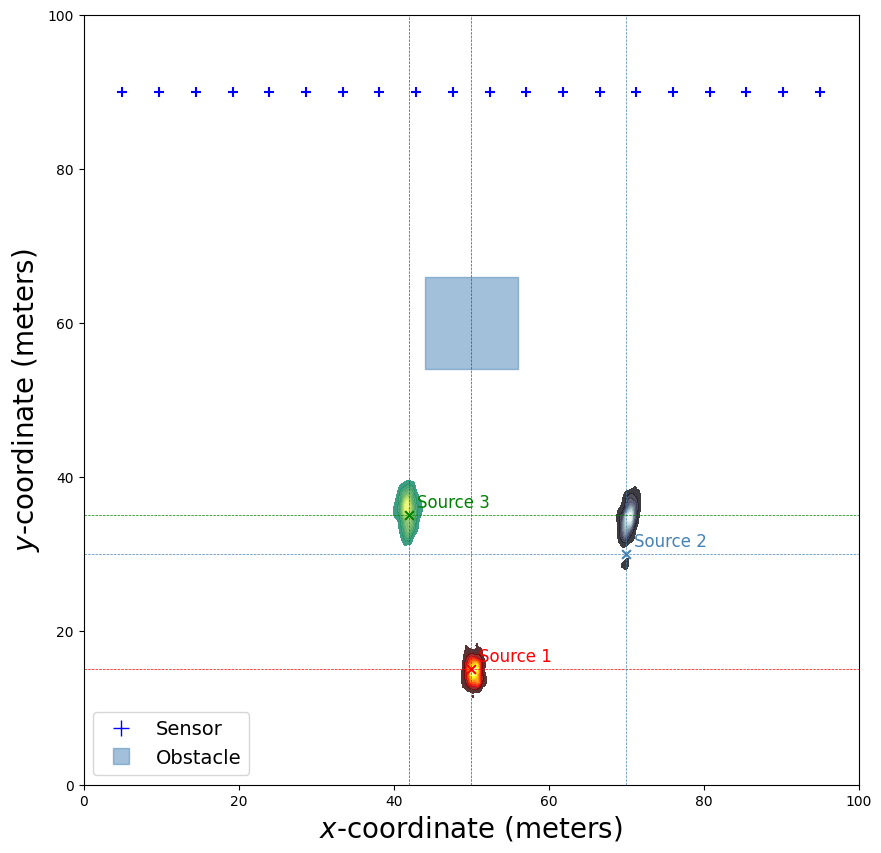}
  \caption{Case study results for surrogate-based SIR filter source location estimation in unsteady and obstructed flow fields. Data is collected using a line of 20 point-sensors. The three location posterior densities show location estimation at the end of the 10 minutes of gas release.}
\label{casestudylocation}
\end{figure}

\section{Conclusion}\label{conclusion}

We have introduced a real-time Bayesian inversion framework for estimating gas emission source location and emission rate in unsteady atmospheric flow fields. By embedding a MLP surrogate for high-fidelity CFD within a sequential importance resampling particle filter, we enable accurate and computationally efficient inference in complex environments. This combination allows for rapid, data-driven updates as new observations arrive, making the method suitable for time-critical spatio-temporal environmental monitoring tasks.

Our experiments demonstrate that the MLP surrogate achieves accuracy comparable to full numerical solvers while requiring orders-of-magnitude less computation. On the Chilbolton dataset, our method outperforms the Gaussian plume model. We further show that the framework generalizes to more challenging, obstructed flow scenarios, accurately recovering hidden sources and tracking temporally varying emission rates.

\paragraph{Limitations and Future Directions.} While our two-dimensional surrogate assumption is appropriate for flat sites with near-sensor level sources, vertical source-sensor offsets could degrade accuracy in more complex terrains. A lightweight extension using vertical Gaussian plume corrections is proposed in the Supplementary Materials. Furthermore, the inference involves the specification of several tuning parameters -- e.g. the number of SIR particles and iterations; and length of the time-window of the data -- appropriate choices for which are likely to be application-dependent. This is also true in how the surrogate model's training data was simulated, some obstacle shapes may obstruct and delay the gas flow requiring CFD runs considerably longer than 1 minute; this is explored in the Supplementary Materials.  Additionally, retraining a separate MLP per time-window introduces overhead, which could be alleviated via online or transfer learning approaches \citep{weiss2016survey}. Finally, robustness to real-world uncertainties -- e.g., sensor dropout, wind misestimation, and multi-source interference -- remains an open avenue for future work. This study underscores the promise of combining physics-informed machine learning with Bayesian inference for scalable, accurate, and fast environmental monitoring in dynamic conditions.


\bibliographystyle{plain}
\bibliography{Ref}

\end{document}